%% file: arxiv-RBC.tex
\documentclass[10pt,twocolumn,letterpaper]{article}

\usepackage[pagenumbers]{cvpr} % To force page numbers, e.g. for an arXiv version

\usepackage{graphicx}
\usepackage{amsmath}
\usepackage{amssymb}
\usepackage{booktabs}
\usepackage{multirow}
\usepackage{arydshln}
\usepackage{latexsym}
\usepackage[pagebackref,breaklinks,colorlinks]{hyperref}

\usepackage[capitalize]{cleveref}
\crefname{section}{Sec.}{Secs.}
\Crefname{section}{Section}{Sections}
\Crefname{table}{Table}{Tables}
\crefname{table}{Tab.}{Tabs.}

\def\cvprPaperID{} % *** Enter the CVPR Paper ID here
\def\confName{CVPR}
\def\confYear{2022}

\begin{document}

%%%%%%%%% TITLE - PLEASE UPDATE
\title{RBC: Rectifying the Biased Context in Continual Semantic Segmentation}

\author{
	Hanbin Zhao$^{1, }$\thanks{The first two authors contributed equally to this paper.}~~~~~
	Fengyu Yang$^{3, }$\footnote[1]{}~~~~~
	Xinghe Fu$^{1}$~~~~~
	Xi Li$^{1, 2, }$\thanks{Corresponding Author}
	\\
	\and
	$^{1}$College of Computer Science and Technology, Zhejiang University\\$^{2}$Shanghai Institute for Advanced Study, Zhejiang University~~~~$^{3}$University of Michigan~~~~
	\and
	{\tt\small 
		\{zhaohanbin, xinghefu, xilizju\}@zju.edu.cn}~~~
	{\tt\small fredyang@umich.edu}
}

\maketitle

%%%%%%%%% ABSTRACT
\begin{abstract}
   Recent years have witnessed a great development of Convolutional Neural Networks in semantic segmentation, where all classes of training images are simultaneously available. In practice, new images are usually made available in a consecutive manner, leading to a problem called Continual Semantic Segmentation (CSS). Typically, CSS faces the forgetting problem since previous training images are unavailable, and the semantic shift problem of the background class. Considering the semantic segmentation as a context-dependent pixel-level classification task, we explore CSS from a new perspective of context analysis in this paper. We observe that the context of old-class pixels in the new images is much more biased on new classes than that in the old images, which can sharply aggravate the old-class forgetting and new-class overfitting. To tackle the obstacle, we propose a biased-context-rectified CSS framework with a context-rectified image-duplet learning scheme and a biased-context-insensitive consistency loss. Furthermore, we propose an adaptive re-weighting class-balanced learning strategy for the biased class distribution. Our approach outperforms state-of-the-art methods by a large margin in existing CSS scenarios.
\end{abstract}

%%%%%%%%% BODY TEXT
\input{sections/introduction.tex}

\input{sections/related.tex}
\input{sections/method.tex}
\input{sections/experiment.tex}

\input{sections/conclusion.tex}

\newpage
%%%%%%%%% REFERENCES
{\small
\bibliographystyle{ieee_fullname}
\bibliography{egbib}
}

\end{document}

%% file: sections/introduction.tex
% !TeX spellcheck = en_US
\section{Introduction}
\label{sec:intro}
Semantic segmentation is a classic pixel-level classification problem in the computer vision area, where deep learning approaches have led to marvelous effect when a large-scale pixel-wise labeled dataset is given~\cite{long2015fully, zhang2018context, chen2017rethinking, yu2018deep, DBLP:conf/cvpr/ChenLLBWWFX020}. However, in a more practical scenario, deep neural networks are required to learn a sequence of tasks with incremental classes and data which is known as the continual learning setup. Semantic segmentation under the setting of continual learning is referred as Continual Semantic Segmentation (CSS)~\cite{cermelli2020modeling, douillard2021plop, michieli2021knowledge}. The study of CSS aims at alleviating the forgetting of the network on past tasks and the overfitting on the current task without past data available.
\begin{figure}[t]
	\begin{center}
		\includegraphics[width=0.9\linewidth]{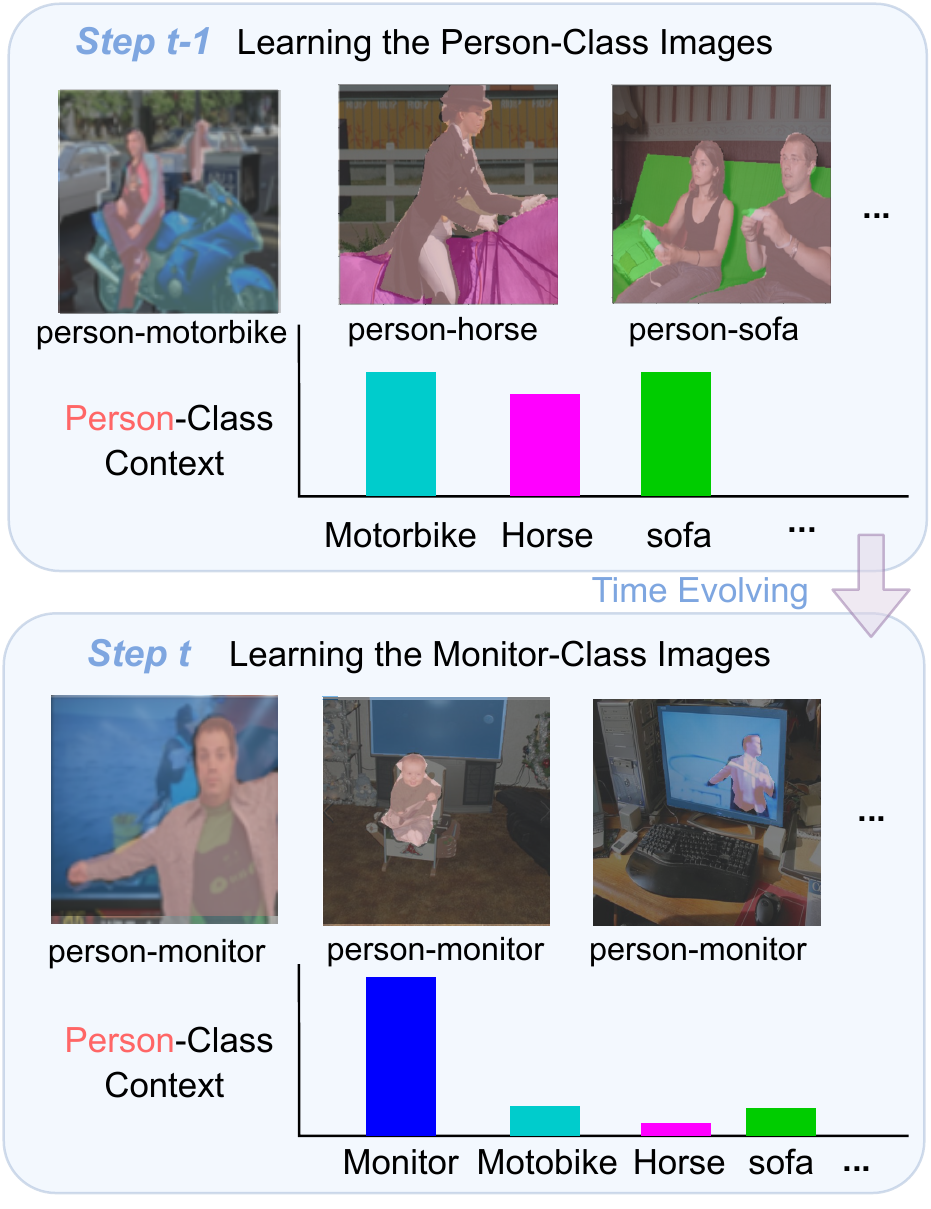}
	\end{center}
	\caption{Illustration of biased context correlation between the old-class and new-class pixels in the added new images. At step $t-1$, the context of \emph{person}-class contains different types (e.g. person-horse, person-sofa) while the model learning person-class images firstly. However, the person context mainly contains person-monitor while the model learning the added \emph{monitor}-class images at step $t$. Thus, a new-class-biased context for the old-class (person) pixels exists in the added new-class (monitor) images, which aggravates the old-class forgetting and new-class overfitting problem of the model.}
	\label{fig:2}
\end{figure}

Currently, there are two main challenges in the study of CSS problem. The first challenge is the catastrophic forgetting phenomenon in continual learning~\cite{mccloskey1989catastrophic}. In CSS, the images for past tasks are usually unavailable while the model learning the current task, and only the pixels belonging to new semantic classes are labeled. The model tends to forget the ability to distinguish pixels belonging to old classes due to the shortage of labeled old-class data in the training stage. The second challenge is CSS-specific and called semantic shift of background class~\cite{cermelli2020modeling}. In the current task of CSS, only new-class pixels are labeled as a semantic class and other pixels including old-class pixels are labeled as background class. This semantic shift of pixel-wise labels causes the ambiguous meaning of old-class pixels during the continual learning process and brings an obstacle to the correct model prediction. 
Since the semantic segmentation is usually considered as a context-dependent pixel-level classification task~\cite{ding2018context}, we explore CSS from the perspective of context. As shown in Figure~\ref{fig:2}, we find out there is another CSS-specific challenge that has not drawn attention. The context of old-class pixels in the new images is much more biased on new classes than that in the old images, which can cause the sharp aggravation of old-class forgetting and new-class overfitting. We call this challenge ``biased context'' in CSS.

\begin{figure}[t]
	\begin{center}
		\includegraphics[width=0.99\linewidth]{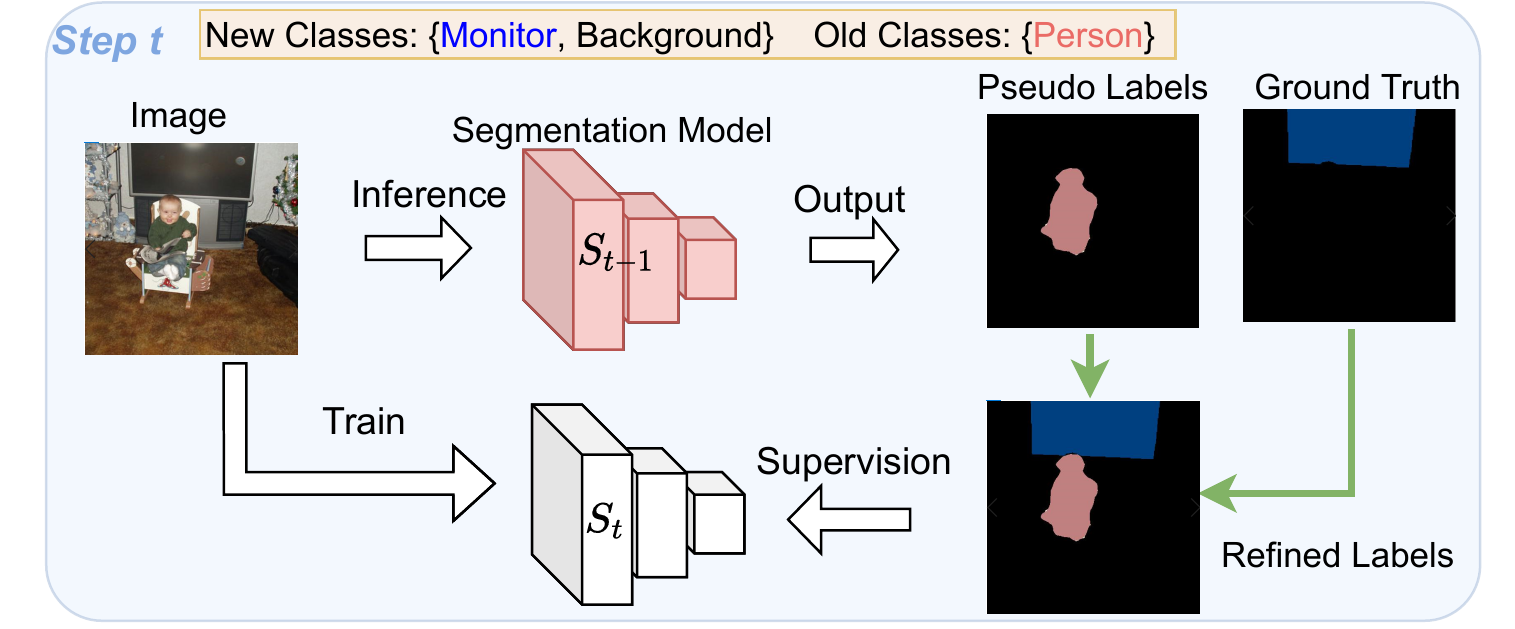}
	\end{center}
	\caption{Pseudo-labeling-based CSS methods. At step $t$, only the new-class (\emph{monitor}) pixels in the added images are labeled and other pixels are all ``background class" pixels. With the old model $S_{t-1}$ from the last step, the mislabeled old-class pixels (\emph{person}) can be pseudo-labeled, and then the model $S_t$ is updated with the old-class pseudo labels and new-class ground truth labels.}
	\label{fig:1}
\end{figure}

In the literature, a number of pseudo-labeling-based CSS methods~\cite{douillard2021plop, yan2021framework} attempt to solve the first two main challenges by labeling the mislabeled pixels of old classes with the model obtained from the last learning step (as shown in Figure~\ref{fig:1}). However, the incrementally updated segmentation model still suffers from the biased prediction towards new classes on account of the following two observations: 1) the new images contain the new-class-biased context for the old-class pixels, and 2) the number of new-class pixels included in the new task is much larger than that of old-class pixels, which is commonly termed as an imbalanced class distribution problem.

Motivated by the observations above, we try to address the CSS from the following two aspects: 1) building a biased-context-rectified CSS learning scheme that is less sensitive to the biased context information of old-class pixels in the incremental images, and 2) developing a class-balance CSS learning strategy for the imbalanced class distribution at different learning steps. We propose a biased-context-insensitive consistency loss, which resorts to a consistency constraint on the context of old classes in an image pair. The duplet of images, consisting of the original image (containing the new-class pixels) and the corresponding erased image (erasing the new-class pixels in the original image), rectify the context of old classes with respect to new classes. Furthermore, we propose an adaptive class-balance CSS learning strategy to cope with the biased class distribution, which adaptively assigns higher weights to the old-class pixels.

Overall, the main contributions of this paper are three-fold: (1) We first consider the biased context in the CSS scenario and propose a biased-context-rectified CSS framework, which aims to avoid overfitting on new classes while not forgetting old classes. (2) We design a novel context-rectified image-duplet learning scheme and a biased-context-insensitive consistency loss that ingeniously rectifies the context of old classes with respect to new classes. To cope with the imbalanced class distribution, we propose an adaptive re-weighting class-balanced learning strategy for CSS. (3) Extensive experiments demonstrate the effectiveness of our method. Our method outperforms several previous CSS approaches by a large margin and obtains state-of-the-art performance.

%% file: sections/related.tex
% !TeX spellcheck = en_US
\section{Related Work}
\noindent\textbf{Continual Learning.} The last years have seen great interest in continual learning (i.e. also called incremental learning or lifelong learning)~\cite{delange2021continual}. Continual learning is first explored on the image classification task with the catastrophic forgetting problem. These works can be categorized into three major families: 1) architectural strategies, 2) rehearsal strategies, 3) regularization strategies. Architectural strategies~\cite{li2019learn, zhao2021and, mallya2018packnet, mallya2018piggyback, aljundi2017expert, abati2020conditional} keep the learned knowledge from previous tasks and acquire new knowledge from the current task by manipulating the network architecture, e.g., parameter masking, network pruning. Rehearsal strategies~\cite{rebuffi2017icarl, hou2019learning, zhao2021memory, wu2019large, liu2020mnemonics} replay old tasks information when learning the new task, and the past knowledge is memorized by storing old tasks’ exemplars or old tasks data distribution via generative models. Regularization strategies~\cite{li2017learning, dhar2019learning} alleviate forgetting by regularization loss terms enabling the updated parameters of networks to retain past knowledge. Continual learning is usually conducted under the task-incremental or the class-incremental learning scenarios. The latter is more challenging because the task identity is unavailable at inference time. 
Recently, continual learning has been also explored on several other computer vision tasks, \emph{e.g.}, incremental object detection~\cite{joseph2021towards}, incremental video classification~\cite{zhao2021video}, incremental instance segmentation~\cite{ganea2021incremental}, continual semantic segmentation~\cite{cermelli2020modeling, douillard2021plop, yan2021framework,michieli2019incremental,fontanel2021detecting,yu2020self, maracani2021recall,huang2021half,zheng2021continual,stan2020unsupervised}. Our work focuses on the CSS problem which can be considered as the class-incremental learning scenario on semantic segmentation. 

\noindent\textbf{Continual Semantic Segmentation.} The forgetting problem in CSS is first considered in ILT~\cite{michieli2019incremental} and the more challenging CSS-tailored problem (background shift) is proposed in MiB~\cite{cermelli2020modeling}. To cope with the problems, some regularization based CSS methods~\cite{douillard2021plop,michieli2021continual} utilizes a confidence-based pseudo-labeling method and a feature-based multi-scale pooling distillation scheme or employs a prototype consistency constraint at the latent space, and some replay-based CSS methods~\cite{maracani2021recall} utilize an extra memory to replay the data for old classes by an extra generative adversarial network or web crawling process. Semantic segmentation is a pixel-wise classification problem~\cite{chen2018encoder,ronneberger2015u,huang2021alignseg,long2015fully,yu2018deep,wang2020deep,pohlen2017full,ji2018semantic,ji2019human} and classifying a local pixel with context information is helpful for reducing the local ambiguities~\cite{zhao2017pyramid,zhang2018context,wang2018non,huang2019ccnet}. Our work first analyzes the effect of biased context in CSS, and we design several biased-context-rectified continual learning strategies tailored for CSS problem. 

%% file: sections/method.tex
% !TeX spellcheck = en_US
\begin{figure*}[t]
	\begin{center}
		\includegraphics[width=0.9\linewidth]{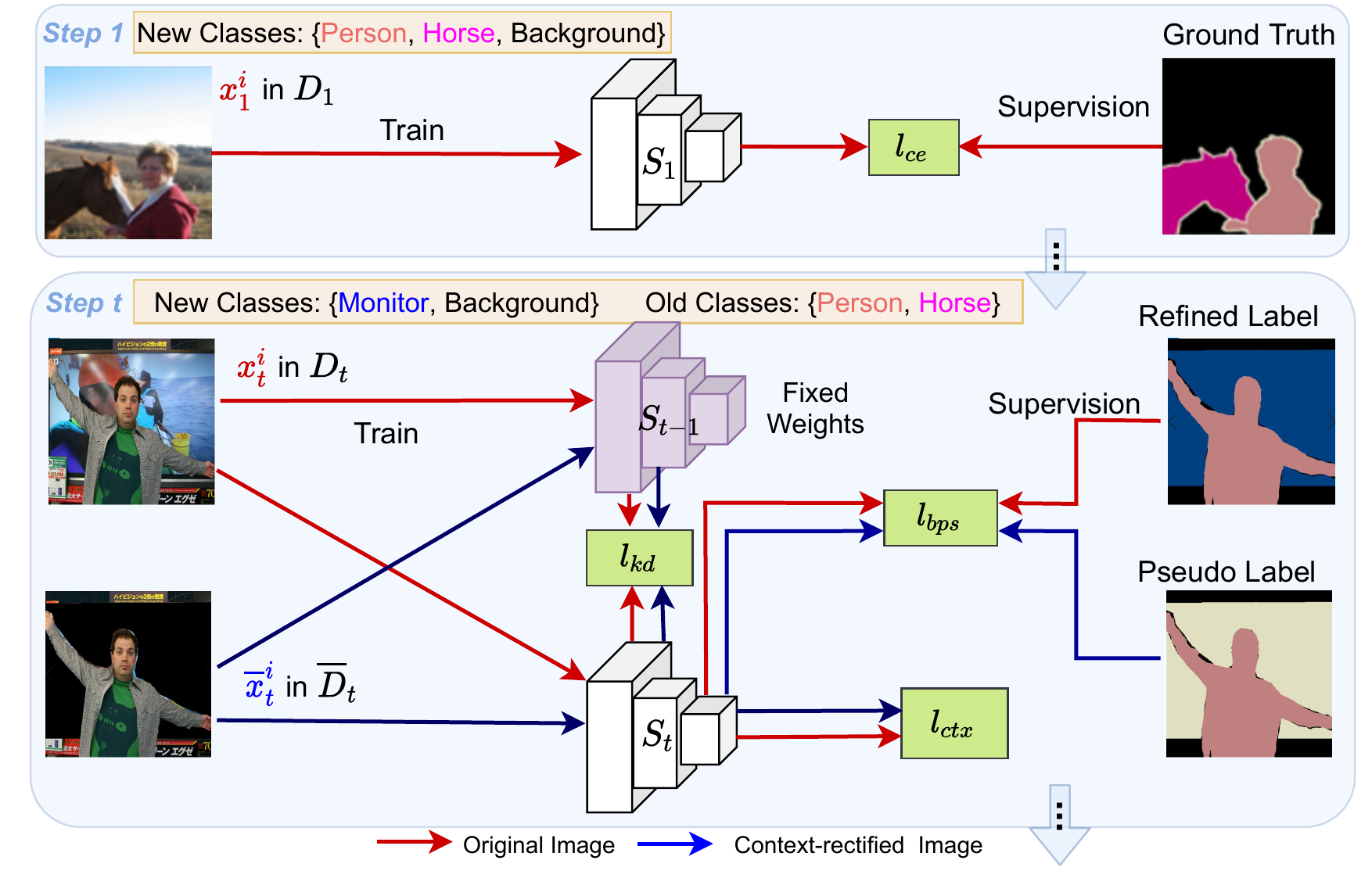}
	\end{center}
	\caption{Illustration of our biased-context-rectified CSS framework. At step 1, the semantic segmentation model is trained from scratch via the classic cross-entropy loss $l_{ce}$ on $D_1$. At the latter steps (e.g. step $t$), we first obtain the context-rectified image-duplet ($D_t,\overline{D}_t$) and update the model by our context-rectified image-duplet learning scheme with the balanced pseudo-labeling loss $l_{bps}$ and the distillation loss $l_{kd}$ and our biased-context-insensitive consistency loss $l_{ctx}$.}
	\label{fig:4}
\end{figure*}
\section{Method}
\subsection{CSS Problem Formulation}
In a continual semantic segmentation scenario, a segmentation model learns several image segmentation tasks continually, and the image subset in each learning step contains pixels from one or several new classes~\cite{cermelli2020modeling, douillard2021plop, michieli2021knowledge}. We suppose the training image set for the $t$-th learning step is $D_{t}$ that consists of a set of pairs $(\mathbf{x}_t^i, \mathbf{y}_t^i)$, where $\mathbf{x}_t^i \in \mathbb{R}^{H\times W\times3}$ and $\mathbf{y}_t^i \in \mathcal{\tilde{Y}}^{H\times W}_t$ denote the $i$-th input image of size $W \times H$ and the corresponding ground truth segmentation mask, respectively. New categories $C_t$ are introduced and required to be learnt at the $t$-th step. $\mathbf{y}_t^i$ only contains the labels of $C_t$ and all other labels (e.g., old classes $C_{1:t-1}$) are collapsed into the background class $C_{0}$. 

We assume a typical semantic segmentation model $S$ with parameters $\Theta$, which consists of an encoder-decoder backbone network $F$ extracting a dense feature map and a convolution head $G$ producing the segmentation score map. Classically, we utilize $S(\mathbf{x})=G\cdot F(\mathbf{x})$ to represent the output predicted segmentation mask of $\mathbf{x}$, $S^{w,h,c}(\mathbf{x})$ denotes the prediction score (about the class $c$) of the pixel at the location $(w,h)$ of $\mathbf{x}$, and $\hat{S}(\mathbf{x})={\rm Softmax}\left(S(\mathbf{x})\right)$ denotes the output of the network. Then, $S_t$ with parameters $\Theta_t$ is updated on $D_t$ at the $t$-th step. Our goal is to obtain $S_t$ which performs well on both previously seen classes $C_{1:t-1}$ and the current classes $C_t$. CSS task is faced with three dilemmas: 1) $S_t$ is only updated on $D_t$ without the previously seen data $D_{1:t-1}$ and suffers from a significant performance drop on pixels of old classes (i.e., the catastrophic forgetting problem); 2) some of pixels in $\mathbf{x}_t^i$ of $D_t$ are mislabeled as $C_{0}$ but actually belong to $C_{1:t-1}$ (i.e., the background shift problem); 3) the context for the old-class pixels in $D_t$ is biased to new classes, since the new-class pixels are usually dominant in the added images $D_t$. 

To address the first two issues, pseudo-labeling CSS methods~\cite{douillard2021plop, yan2021framework} are proposed by labeling the mislabeled pixels with the model obtained from the last step, which is described in Section~\ref{section_3.2}. These methods alleviate the forgetting problem since a few pixels of old classes are introduced during learning the new images (similar to the replay-based continual learning strategy~\cite{rebuffi2017icarl, wu2019large, hou2019learning}), and reduce the background shift due to correcting the mislabeled ``background class" pixels. However, the updated segmentation model by these methods still suffers from the biased prediction towards new classes because of the following two observations: the biased context (shown in Figure~\ref{fig:2}) and the common imbalanced class distribution in the new images. To alleviate the above issues, we propose a biased-context-rectified CSS framework including a context-rectified image-duplet learning scheme and a biased-context-insensitive consistency loss in Section~\ref{section_3.3} and the illustration of our framework is shown in Figure~\ref{fig:4} and propose an adaptive class-balance strategy for tackle the biased class distribution in Section~\ref{section_3.4}.

\subsection{Pseudo-Labeling-Based CSS}\label{section_3.2}
To alleviate the forgetting and background shift problems, pseudo-labeling-based methods~\cite{douillard2021plop, yan2021framework} are utilized in CSS.
Specifically, at the $t$-th learning step, we can access to $S_{t-1}$ from the last step and correct the mislabeled ``background class" pixels with $S_{t-1}$ (as shown in Figure~\ref{fig:1}). For each $(\mathbf{x}_t^i, \mathbf{y}_t^i)$ in $D_t$, the pixels belonging to the new classes $C_t$ have ground-truth labels and some of the other pixels belonging to the old classes $C_{1:t-1}$ are mislabeled as $C_{0}$. The predictions of the old model for these mislabeled pixels $\hat{S}_{t-1}(\mathbf{x}_t^i)$ are utilized as clues if they belong to any of the old classes. After that, each $\mathbf{x}_t^i$ in $D_t$ can have a refined segmentation label $\tilde{S}_{t}(\mathbf{x}_t^i)$ by combining the pseudo label $\hat{S}_{t-1}(\mathbf{x}_t^i)$ and the ground truth $\mathbf{y}_t^i$ (as shown in Figure~\ref{fig:1}). Then the model $S_t$ is updated by optimizing the following objective function:
\begin{equation}\label{eq_1}
	\begin{aligned}
		\mathcal{L}_{total}(\Theta_t) = \frac{1}{\left|D_t\right|}\sum_{(\mathbf{x},\mathbf{y})\in D_t}l(\mathbf{x};\Theta_t),
	\end{aligned}
\end{equation}
where $l(\mathbf{x};\Theta_t)$ is usually composed of a cross-entropy loss term with pseudo-labeling and a knowledge distillation term:
\begin{equation}\label{eq_1-2}
\begin{aligned}
l(\mathbf{x};\Theta_t) = l_{ps}(\mathbf{x};\Theta_t) + \alpha l_{kd}(\mathbf{x};\Theta_t),
\end{aligned}
\end{equation}
where $\alpha$ is a hyper-parameter balancing the importance of the loss terms. $l_{ps}(\mathbf{x};\Theta_t)$ is utilized to maintain the performance on old classes and reduce the ambiguity of old-class pixels labeled as background class at step $t$:
\begin{equation}\label{eq_2}
\begin{aligned}
l_{ps}(\mathbf{x};\Theta_t) = -\frac{\beta}{WH}\sum_{w,h}^{W,H}\sum_{c\in C_{0:t} }\tilde{S}^{w,h,c}_{t}(\mathbf{x})\log \hat{S}^{w,h,c}_{t}(\mathbf{x}),
\end{aligned}
\end{equation}
where $\beta$ is the ratio of accepted old classes pixels over the total number of such pixels. $l_{kd}(\mathbf{x};\Theta_t)$ is added to the backbone network $F(\cdot)$ to retain information of the old classes:
\begin{equation}\label{eq_3}
\begin{aligned}
l_{kd}(\mathbf{x};\Theta_t) = \left\|\Phi(F_{t}(\mathbf{x})))-\Phi(F_{t-1}(\mathbf{x})))\right\|^2,
\end{aligned}
\end{equation}
where $\left\|\cdot \right\|$ and $\Phi(F(\mathbf{x}))\in \mathbb{R}^{(H+W)\times C}$ denotes the Euclidean distance and concatenation operation, respectively. The concatenation operation function $\Phi(F(\mathbf{x}))$ is formulated as follows:
\begin{equation}\label{eq_3-2}
\begin{aligned}
\Phi(F(\mathbf{x}))=\left[\frac{1}{W}\sum_{w=1}^{W}F^{:,w,:}(\mathbf{x})\big|\big|\frac{1}{H}\sum_{h=1}^{H}F^{h,:,:}(\mathbf{x})\right],
\end{aligned}
\end{equation}
where $\left[\cdot||\cdot\right]$ denotes concatenation over the channel axis.

\subsection{Biased-context-rectified framework}\label{section_3.3}
To alleviate the biased context correlation between the old-class and new-class pixels in CSS, we propose a biased-context-rectified framework with a context-rectified image-duplet learning scheme and a biased-context-insensitive consistency loss. Taking the $t$-th step as an example (shown in Figure~\ref{fig:2}), the incrementally added images $D_t$ mainly contain the new-class-related context for the old-class pixels, which leads to the aggravation of the old-class forgetting and new-class overfitting problems.

\begin{figure}[t]
	\begin{center}
		\includegraphics[width=0.99\linewidth]{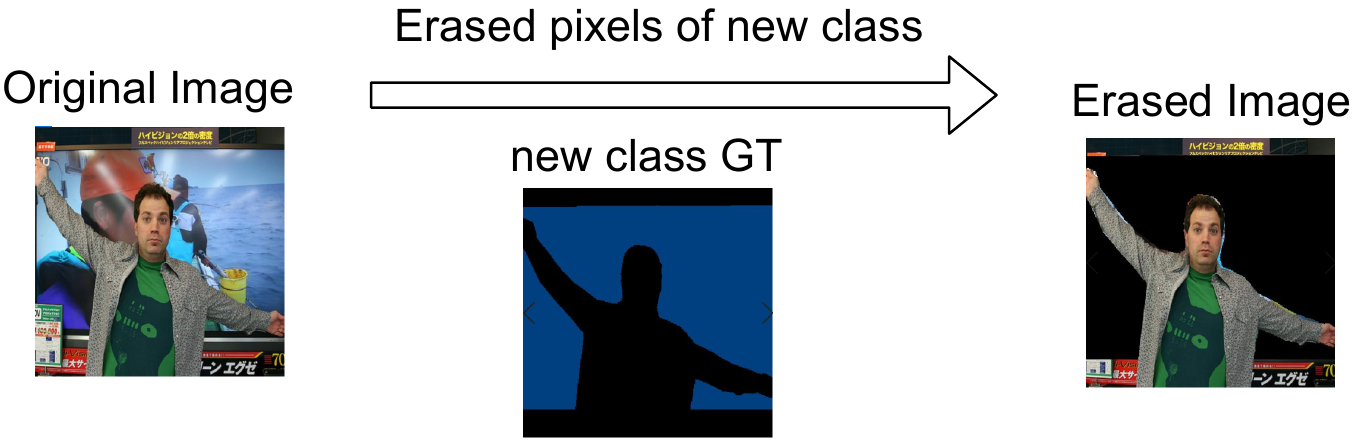}
	\end{center}
	\caption{The illustration of generating the context-rectified image-duplet.}
	\label{fig:5}
\end{figure}

\noindent\textbf{Context-rectified Image-Duplet Learning.}
As for the new-class-related context, we observe that the contextual information of old-class pixels included in the incremental images is biased to the pixels of new classes (shown in Figure~\ref{fig:2}). In order to continually learn a semantic segmentation model that is less sensitive to the entangled new-class-context, we firstly rectify the biased context between new classes and old classes in these new images by erasing the new-class pixels of the original image (shown in Figure~\ref{fig:5}). At the $t$-th step, we obtain the corresponding erased image $\mathbf{\overline{x}}_t^i$ for each new image $\mathbf{x}_t^i$ in $D_t$.
Then an image-duplet $(\mathbf{x}_t^i, \mathbf{y}_t^i, \mathbf{\overline{x}}_t^i, \mathbf{\overline{y}}_t^i)$ is constructed from the erased image and the corresponding original image. The set of image-duplets with $D_t$ and $\overline{D}_t$ are denoted as:
\begin{equation}\label{eq_4}
\begin{aligned}
&(D_t,\overline{D}_t) = \left\{(\mathbf{x}_t^i, \mathbf{y}_t^i, \mathbf{\overline{x}}_t^i, \mathbf{\overline{y}}_t^i)\right\}_{i=1}^{\left|D_t\right|}\\
&s.t.~(\mathbf{x}_t^i, \mathbf{y}_t^i) \in D_t, (\mathbf{\overline{x}}_t^i, \mathbf{\overline{y}}_t^i) \in \overline{D}_t,\\
\end{aligned}
\end{equation}
 
 With the image-duplets $(D_t,\overline{D}_t)$ at the $t$-th step, our method updates the model $S_t$ by optimizing the following loss function:
\begin{equation}\label{eq_5-1}
\begin{aligned}
\mathcal{L}_{total}(\Theta_{\scriptscriptstyle t}) =& \frac{1}{\scriptscriptstyle \left|D_t\right|+\left|\overline{D}_t\right|}\sum_{\scriptscriptstyle (\mathbf{x},\mathbf{\overline{x}})}[l_{\scriptscriptstyle dup}(\mathbf{x},\mathbf{\overline{x}};\Theta_t)
+\gamma l_{\scriptscriptstyle ctx}(\mathbf{x},\mathbf{\overline{x}};\Theta_{\scriptscriptstyle t})],\\
\end{aligned}
\end{equation}
where $\gamma$ is a hyper-parameter balancing the importance of the loss terms. The first loss term $l_{dup}(\mathbf{x},\mathbf{\overline{x}};\Theta_t)$ takes the similar form of Equation~\eqref{eq_1-2} on the original image $\mathbf{x}$ and the corresponding erased image $\mathbf{\overline{x}}$:
 \begin{equation}\label{eq_5-2}
\begin{aligned}
l_{dup}(\mathbf{x},\mathbf{\overline{x}};\Theta_t) = l(\mathbf{x};\Theta_t) + l(\mathbf{\overline{x}};\Theta_t),
\end{aligned}
\end{equation} 
   
\begin{figure}[t]
	\begin{center}
		\includegraphics[width=0.99\linewidth]{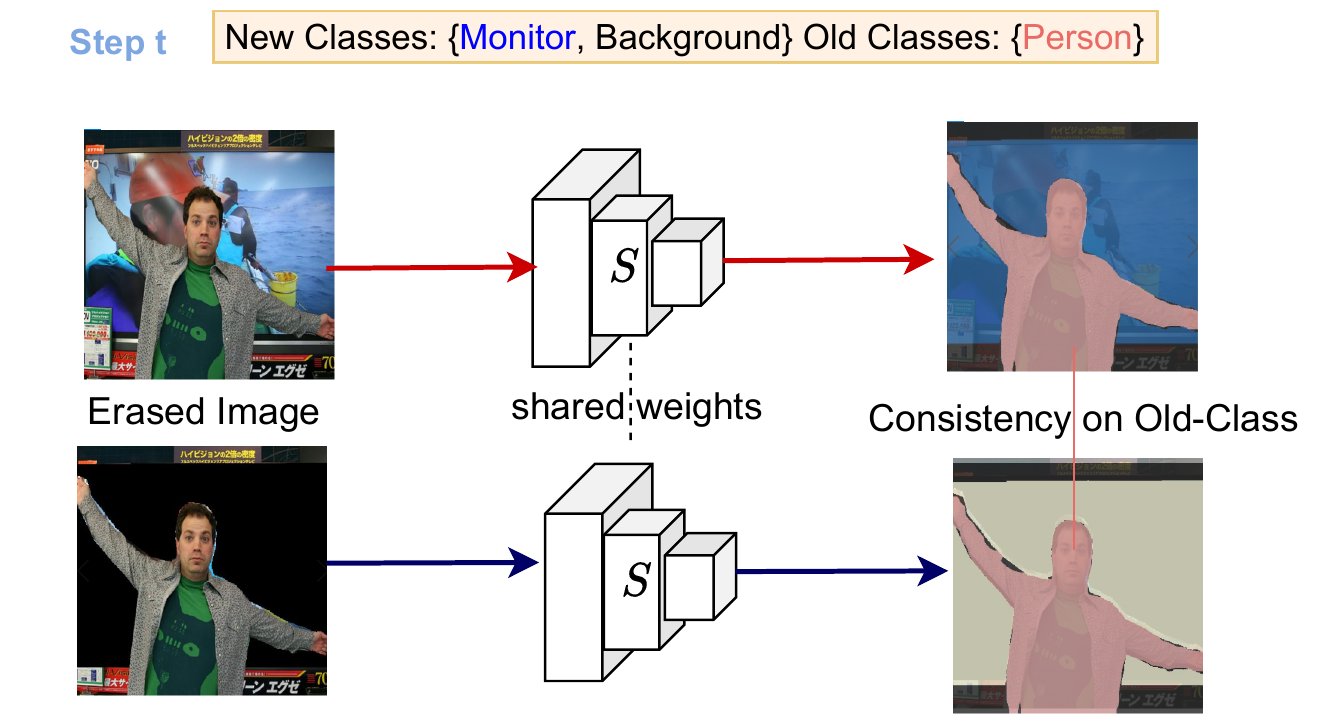}
	\end{center}
	\caption{The illustration of a biased-context-insensitive consistency loss.}
	\label{fig:6}
\end{figure}

\noindent\textbf{Biased-context-insensitive Consistency Loss.}
 To further address the biased context, the second loss term $l_{ctx}(\mathbf{x},\mathbf{\overline{x}};\Theta_t)$ is introduced and utilized to keep a biased-context-insensitive consistency between the original image $\mathbf{x}$ and the corresponding erased image $\mathbf{\overline{x}}$ (as shown in Figure~\ref{fig:6}). For the old-class pixels, the new-class-related context are included in the original image $\mathbf{x}$ and erased in the corresponding erased image $\mathbf{\overline{x}}$. For simplicity, we utilize $O(\mathbf{x})$ to represent the locations $\{(w^{j}_o,h^{j}_o)\}^{O(\mathbf{x})}_{j=1}$ of old-class pixels included in the image $\mathbf{x}$, .
 To reduce the effect of biased context between the old-class and new-class, the prediction of the updated model $S_t$ on the old-class pixels with the new-class-related context should be consistent with that without the new-class-related context.
 Then $l_{ctx}(\mathbf{x},\mathbf{\overline{x}};\Theta_t)$ is formulated as follows:
 \begin{equation}\label{eq_5-3}
 \begin{aligned}
 &l_{\scriptscriptstyle ctx}(\mathbf{x},\mathbf{\overline{x}};\Theta_{\scriptscriptstyle t}) = \\
 &\sum_{\scriptscriptstyle (w,h) \in O(\mathbf{x})}^{}\sum_{\scriptscriptstyle c\in C_{1:t-1} }\left\|S^{\scriptscriptstyle w,h,c}_{\scriptscriptstyle t}(\mathbf{\overline{x}})-S^{\scriptscriptstyle w,h,c}_{\scriptscriptstyle t}(\mathbf{x})\right\|^2,\\
 \end{aligned}
 \end{equation}

\subsection{Adaptive Class-Balance CSS}\label{section_3.4}
 As for the imbalanced class distribution problem, we observe that the number of new-class pixels included in the new images is much larger than that of pseudo-labeled old-class pixels. This class-imbalance problem usually results in the updated classifier being biased towards the new classes. To cope with the problem, we propose to adaptively assign different weights to the pixels of different classes based on the number of pixels. We optimize the biased classifier by a balanced pseudo-labeling cross-entropy loss $l_{bps}(\mathbf{x};\Theta_t)$ with different weights.

To address the class-imbalance problem in CSS, the balanced pseudo-labeling cross-entropy loss is formulated as follows:
\begin{equation}\label{eq_7}
\begin{aligned}
l_{ \scriptscriptstyle bps}(\mathbf{x};\Theta_{\scriptscriptstyle t}) = -\frac{\scriptscriptstyle \beta}{\scriptscriptstyle WH}\sum_{\scriptscriptstyle w,h}^{\scriptscriptstyle W,H}\sum_{c\in C_{\scriptscriptstyle 0:t} }\eta^{w,h}(\mathbf{x})\tilde{S}^{\scriptscriptstyle w,h,c}_{ \scriptscriptstyle t}(\mathbf{x})\log \hat{S}^{\scriptscriptstyle w,h,c}_{\scriptscriptstyle t}(\mathbf{x}),
\end{aligned}
\end{equation}
where $\eta^{w,h}(\mathbf{x})$ denotes the weight of the pixel at the location $(w,h)$ in the image $\mathbf{x}$. $\eta^{w,h}(\mathbf{x})$ depends on the category of the pixel and the number of pixels from different classes in the image:
\begin{equation}\label{eq_8}
\begin{aligned}
\eta^{w,h}(\mathbf{x}) =
\begin{cases} 
0.5+\sigma(\frac{N^{old}(\mathbf{x})}{N^{new}(\mathbf{x})})&  \scriptstyle (w,h) \in O(\mathbf{x})\\
1  &  {\rm otherwise},
\end{cases}
\end{aligned}
\end{equation}
where $N^{old}(\mathbf{x})$, $N^{new}(\mathbf{x})$ and $\sigma(\cdot)$ are the number of pixels belonging to old classes $C_{1:t-1}$, the total number of pixels belonging to the new classes $C_t$ and the sigmoid function respectively. Then $l_{dup}(\mathbf{x},\mathbf{\overline{x}};\Theta_t)$ in Equation~\eqref{eq_5-2} with the balanced pseudo-labeling is formulated as follows: 
\begin{equation}\label{eq_9}
\begin{aligned}
l_{dup}(\mathbf{x},\mathbf{\overline{x}};\Theta_t)= l^{'}(\mathbf{x};\Theta_t) + l(\mathbf{\overline{x}};\Theta_t),
\end{aligned}
\end{equation}
where $l^{'}(\mathbf{x};\Theta_t)$ is denoted as follows:
\begin{equation}\label{eq_9-2}
\begin{aligned}
l^{'}(\mathbf{x};\Theta_t)= l_{bps}(\mathbf{x};\Theta_t) + \alpha l_{kd}(\mathbf{x};\Theta_t),
\end{aligned}
\end{equation}

%% file: sections/experiment.tex
% !TeX spellcheck = en_US
\section{Experiments}
\subsection{Experimental Setup}
\noindent\textbf{Datasets:} we follow previous CSS works~\cite{maracani2021recall,cermelli2020modeling,douillard2021plop} and utilize the commonly used semantic segmentation dataset PASCAL VOC 2012~\cite{everingham2010pascal} for experiments: it contains $10,582$ fully-annotated images for training and $1,449$ for testing, over $20$ foreground object classes. For the dataset, we resize the images to $512\times 512$, with a center crop and employ the random horizontal flip augmentation strategy as the practice in PLOP~\cite{douillard2021plop} at training time.

\noindent\textbf{Evaluation Protocol:} MiB~\cite{cermelli2020modeling} introduces two different CSS settings (\emph{Disjoint} and \emph{Overlapped}). In the \emph{Disjoint} setting, the incremental new images $D_t$ at $t$-th step contain pixels belonging to old and current new classes ($C_{1:t-1}\cup C_{t}$). In the \emph{Overlapped} setting, the new images contains the pixels belonging to old, current new and future classes ($C_{1:t-1}\cup C_{t} \cup C_{t+1:T}$). The \emph{Overlapped} setting is usually more challenging than the \emph{Disjoint} setting.
We evaluate our method under the above two settings on three commonly used CSS benchmarks (VOC-19-1, VOC-15-5, and VOC-15-1), where 19-1 means learning 19 then 1 class (2 learning steps), 15-5 learning 15 then 5 classes (2 steps) and 15-1 learning 15 classes followed by five times 1 class (6 steps). 
The protocol with a higher number of steps is usually more challenging. 
Each method is trained on the CSS benchmark in several steps. At the last step, we follow~\cite{douillard2021plop,michieli2021continual} and report the traditional mean Intersection over Union (mIoU) for the initial classes $C_1$, for the incremented classes $C_{2:T}$, for all classes $C_{1:T}$ (\emph{all}).

\noindent\textbf{Training Details:}
We implement our models with Pytorch and use SGD for optimization. Following \cite{cermelli2020modeling, douillard2021plop}, we use the Deeplab-V3~\cite{chen2017rethinking} architecture with a ResNet-101~\cite{he2016deep} pre-trained on ImageNet~\cite{deng2009imagenet} as the backbone network. Our framework is implemented based on the code of pseudo-labeling-based CSS method PLOP\footnote{https://github.com/arthurdouillard/CVPR2021\_PLOP}~\cite{douillard2021plop} and utilize the same parameters setting as default (e.g. the number of training epochs, the learning rates, the decay rate, the batch size). As for our proposed context-rectified image-duplet learning scheme, we train our model with a batch size of $24$ on Pascal VOC, respectively. At the first CSS step, all the images are the original images since no old model is kept for pseudo-labeling. At other CSS steps, the sample duplets are generated by the old model from the last step (half of the images in each batch are the original images and half of them are the corresponding new-class-erased images). The loss weight $\gamma$ of the biased-context-insensitive consistency loss term in Equation~\eqref{eq_5-1} is set to $0.01$ for all datasets.

\begin{table}[t]
	\centering
	\caption{Ablation experimental results on VOC-\emph{Overlapped}-15-1.}
	\resizebox{0.49\textwidth}{!}{
		\begin{tabular}{llcccc}
			\toprule
			\multirow{2}{*}{\textbf{Ablation}}&\multirow{2}{*}{\textbf{Method}}&\multicolumn{3}{c}{\textbf{15-1} (6 steps)}\\
			\cline{3-6}
			&&0-15&16-20&\emph{all}\\
			\midrule
			\multirow{3}{*}{\emph{Ablation I}}&Baseline
			&65.12&21.11&54.64\\
			&Baseline+double
			&60.23&11.95&48.73\\
			&Baseline+duplet
			&\textbf{70.54}&\textbf{31.06}&\textbf{61.14}\\
			\midrule
			\multirow{2}{*}{\emph{Ablation II}}&Baseline+duplet
			&\textbf{70.54}&31.06&61.14\\
			&Baseline+duplet+ctx
			&69.54&\textbf{38.44}&\textbf{62.14}\\
			\midrule
			\multirow{2}{*}{\emph{Ablation III}}&Baseline
			&65.12&21.11&54.64\\
			&Baseline+balance
			&\textbf{65.35}&\textbf{24.89}&\textbf{55.72}\\
			\bottomrule
		\end{tabular}
	}
	\label{tab:Ab-1}
\end{table}

\begin{figure}[t]
	\begin{center}
		\includegraphics[width=0.95\linewidth]{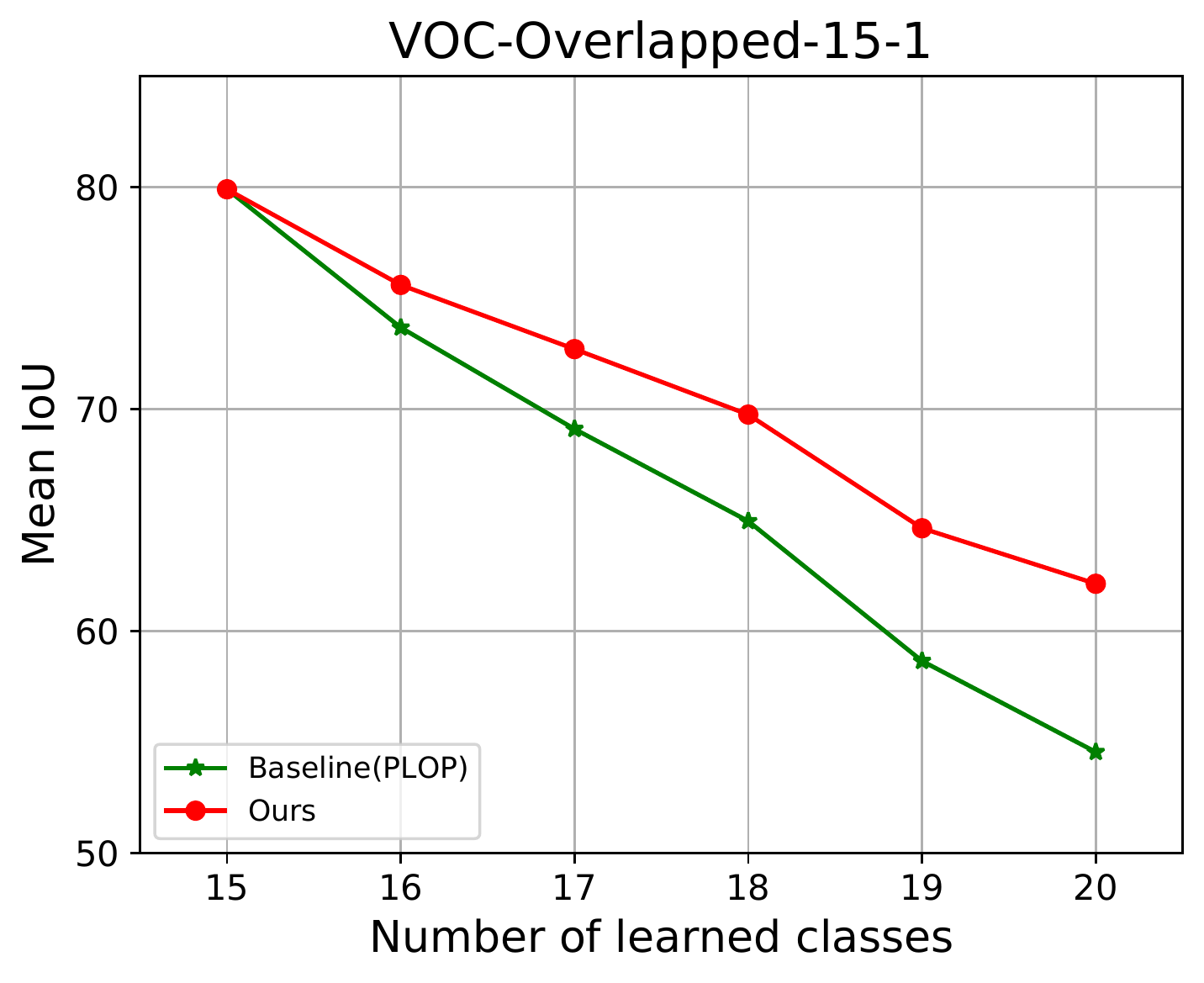}
	\end{center}
	\caption{The mIoU evolution of ours and Baseline (PLOP) on VOC-\emph{Overlapped}-15-1.}
	\label{fig:8}
\end{figure}

\begin{figure*}[t]
	\begin{center}
		\includegraphics[width=0.95\linewidth]{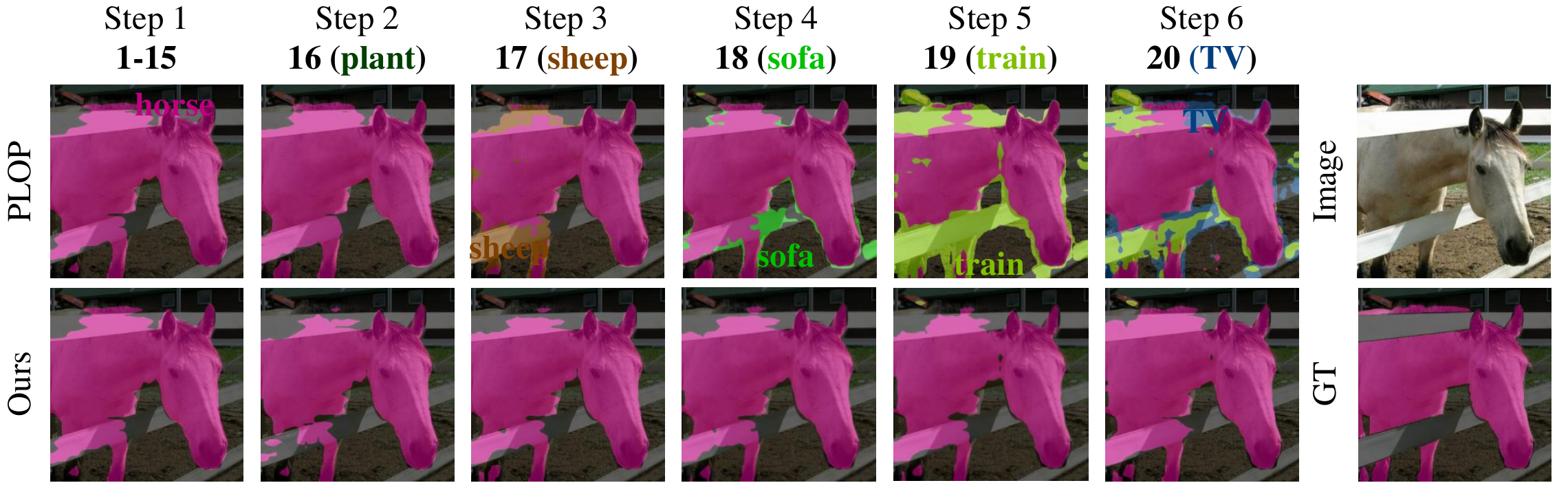}
	\end{center}
	\caption{Visualization of PLOP and our method at different steps under the \emph{Overlapped} setting of VOC-15-1.}
	\label{fig:7}
	\vspace{-3pt}
\end{figure*}

\begin{figure}[t]
	\begin{center}
		\includegraphics[width=1\linewidth]{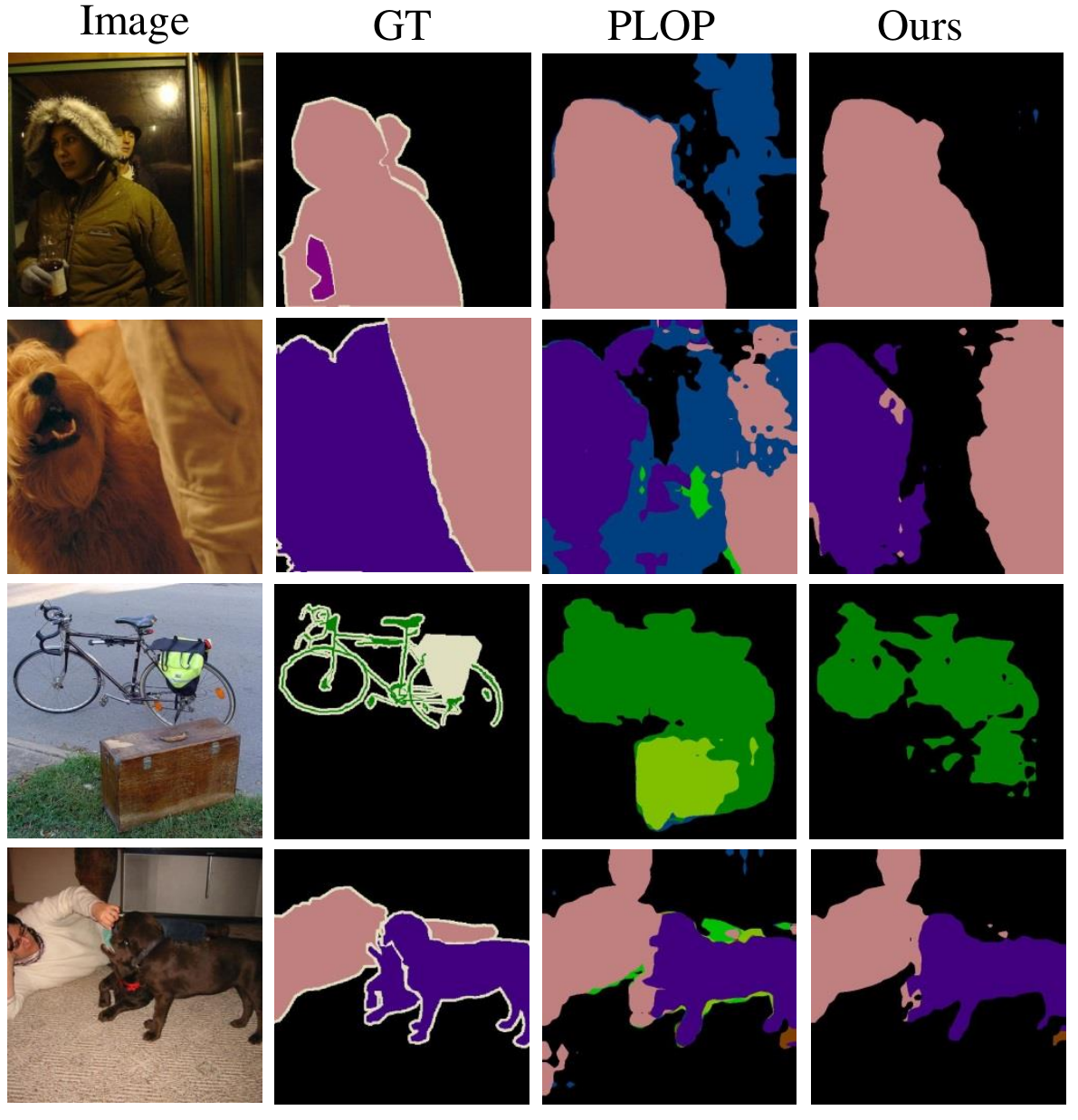}
	\end{center}
	\caption{The predictions of PLOP and our method for different images at the last step under the \emph{Overlapped} setting of VOC-15-1.}
	\label{fig:7-2}
	\vspace{-4pt}
\end{figure}

\subsection{Ablation Study}
In this section, we first carry out ablation experiments to validate the effectiveness of the image duplet. Then we conduct experiments to validate our context-rectified image-duplet learning scheme, the biased-context-insensitive consistency loss and adaptive class-balance strategy.
All of the ablation experiments are conducted on the challenging \emph{Overlapped} setting of the benchmark VOC-15-1.

\noindent\textbf{Baseline.} Our main baseline is given based on a classical pseudo-labeling-based CSS method PLOP~\cite{douillard2021plop}, which utilizes a multi-scale pooling distillation scheme to preserve the performance on previously seen classes and an entropy-based pseudo-labeling strategy on the mislabeled background class pixels to reduce the background shift.

\noindent\textbf{Effect of the context-rectified image-duplet.} In order to demonstrate the effectiveness of the context-rectified image-duplet, we compare the performance of ``Baseline" with our image-duplet including the original image and the corresponding new-class-erased image (denoted by ``Baseline+duplet"). The results are shown in Table~\ref{tab:Ab-1}. We can see that the performance of Baseline+duplet surpasses that of Baseline. In particular, the mIoU on new classes of Baseline+duplet outperforms that of Baseline by a large margin ($9.95\%$). Moreover, we also compare Baseline+duplet with Baseline+double to reduce the influence of increasing the number of samples. In a minibatch, Baseline+duplet utilizes the original images and the corresponding erased images. Baseline+double utilizes the original images and the corresponding copied original images. As shown in Table~\ref{tab:Ab-1}, the performance of Baseline+duplet is higher than Baseline+double, which demonstrates that directly increasing the number of images can not lead to performance improvement.

\noindent\textbf{Effect of biased-context-insensitive consistency constraint.} We evaluate the performance of ``Baseline+duplet" with our biased-context-insensitive consistency loss (denoted by ``Baseline+duplet+ctx") and Table~\ref{tab:Ab-1} summarizes the experimental results on the \emph{Overlapped} setting of the benchmark VOC-15-1. The mIoU on new classes of Baseline+duplet+ctx is around $7.38\%$ higher than that of Baseline+duplet, which demonstrates that the biased-context-insensitive consistency constraint can greatly improve the performance on new classes and is essential to avoid the overfitting on new classes. To further validate the effectiveness of our biased-context-rectified CSS learning framework, we also compare our method with Baseline and show the average mIoU curves in Figure~\ref{fig:8}. It is observed that Ours achieves better performance than Baseline at every learning step.

\noindent\textbf{Effect of adaptive class-balance strategy.} To demonstrate the effectiveness of our adaptive class-balance strategy, we evaluate the performance of ``Baseline" with our adaptive class-balance strategy (denoted by ``Baseline+balance"). In Table~\ref{tab:Ab-1}, we report the experimental results after the last learning step. As for the old classes (0-15), Baseline+balance achieves better performance than Baseline. Regarding new classes (16-20), Baseline+balance exceeds Baseline by around $4\%$. 

\noindent\textbf{Effect of biased context.} To demonstrate the effect of the biased context, we visualize the predictions for both PLOP (Baseline) and our method on 15-1 protocol of the benchmark VOC-\emph{Overlapped}. As shown in Figure~\ref{fig:7}, PLOP is more prone to overfitting on new classes (sheep, sofa, train, TV) than ours at the latter steps. Besides, we visualize the predictions of ours and PLOP for different samples at the last step in Figure~\ref{fig:7-2}. Ours achieve less forgetting on old classes (person, dog, bicycle) than PLOP, illustrating that the biased context aggravates the old-class forgetting and new-class overfitting.

\begin{table*}[t]
	\centering
	\caption{CSS results under the \emph{Disjoint} and \emph{Overlapped} settings of VOC-19-1, VOC-15-5 and VOC-15-1 benchmarks. $\dagger$ means the results from~\cite{douillard2021plop}, and $*$ means the results from~\cite{michieli2021continual}. Best in \textbf{bold}.}
	\resizebox{0.9\textwidth}{!}{
		\begin{tabular}{llccccccccccc}
			\toprule
			\multirow{2}{*}{\textbf{Setting}}&\multirow{2}{*}{\textbf{Method}}&\multicolumn{3}{c}{\textbf{19-1} (2 steps)}&&\multicolumn{3}{c}{\textbf{15-5} (2 steps)}&&\multicolumn{3}{c}{\textbf{15-1} (6 steps)}\\
			\cline{3-5}\cline{7-9}\cline{11-13}
			&&0-19&20&\emph{all}&
			&0-15&16-20&\emph{all}&
			&0-15&16-20&\emph{all}\\
			\midrule
			\multirow{11}{*}{\emph{Disjoint}}&FT&5.80&12.30&6.20&
			&1.10&33.60&9.20&
			&0.20&1.80&0.60\\
			\cline{2-13}
			&PI$^\dagger$~\cite{zenke2017continual} &5.40&14.10&5.90&
			&1.30&34.10&9.50&
			&0.00&1.80&0.40\\
			&EWC$^\dagger$~\cite{kirkpatrick2017overcoming} &23.20&16.00&22.90&
			&26.70&37.70&29.40&
			&0.30&4.30&1.30\\
			&RW$^\dagger$~\cite{chaudhry2018riemannian} &19.40&15.70&19.20&
			&17.90&36.90&22.70&
			&0.80&3.60&1.30\\
			&LwF$^\dagger$~\cite{li2017learning} &53.00&9.10&50.80&
			&58.40&37.40&53.10&
			&0.80&3.60&1.50\\
			&LwF-MC$^\dagger$~\cite{rebuffi2017icarl} &63.00&13.20&60.50&
			&67.20&41.20&60.70&
			&4.50&7.00&5.20\\
			&ILT$^\dagger$~\cite{michieli2019incremental} &69.10&16.40&66.40&
			&63.20&39.50&57.30&
			&3.70&5.70&4.20\\
			&MiB$^\dagger$~\cite{cermelli2020modeling} &69.60&25.60&67.40&
			&71.80&43.30&64.70&
			&46.20&12.90&37.90\\
			&PLOP$^\dagger$~\cite{douillard2021plop} &75.37&38.89&73.64&
			&71.00&42.82&64.29&
			&57.86&13.67&46.48\\  
			\cdashline{2-13}[1pt/1pt]
			&	Ours &\textbf{76.43}&\textbf{45.79}&\textbf{75.01}&
			&\textbf{75.12}&\textbf{49.71}&\textbf{69.89}&
			&\textbf{61.68}&\textbf{19.52}&\textbf{51.60}\\
			\cline{2-13}
			&	Joint&77.40&78.00&77.40&
			&79.10&72.56&77.39&
			&79.10&72.56&77.39\\
			\midrule
			\multirow{11}{*}{\emph{Overlapped}}&FT&6.80&12.90&7.10&
			&2.10&33.10&9.80&
			&0.20&1.80&0.60\\
			\cline{2-13}
			&PI$^\dagger$~\cite{zenke2017continual} &7.50&14.00&7.80&
			&1.60&33.30&9.50&
			&0.00&1.80&0.50\\
			&EWC$^\dagger$~\cite{kirkpatrick2017overcoming} &26.90&14.00&26.30&
			&24.30&35.50&27.10&
			&0.30&4.30&1.30\\
			&RW$^\dagger$~\cite{chaudhry2018riemannian} &23.30&14.20&22.90&
			&16.60&34.90&21.20&
			&0.00&5.20&1.30\\
			&LwF$^\dagger$~\cite{li2017learning} &51.20&8.50&49.10&
			&58.90&36.60&53.30&
			&1.00&3.90&1.80\\
			&LwF-MC$^\dagger$~\cite{rebuffi2017icarl} &64.40&13.30&61.90&
			&58.10&35.00&52.30&
			&6.4&8.40&6.90\\
			&ILT$^\dagger$~\cite{michieli2019incremental} &67.75&10.88&65.05&
			&67.08&39.23&60.45&
			&8.75&7.99&8.56\\
			&MiB$^\dagger$~\cite{cermelli2020modeling} &71.43&23.59&69.15&
			&76.37&49.97&70.08&
			&34.22&13.50&29.29\\
			&PLOP$^\dagger$~\cite{douillard2021plop} &75.35&37.35&73.54&
			&75.73&51.71&70.09&
			&65.12&21.11&54.64\\
			\cdashline{2-13}[1pt/1pt]
			&Ours &\textbf{77.26}&\textbf{55.60}&\textbf{76.23}&
			&\textbf{76.59}&\textbf{52.78}&\textbf{70.92}&
			&\textbf{69.54}&\textbf{38.44}&\textbf{62.14}\\
			\cline{2-13}
			&Joint&77.40&78.00&77.40&
			&79.10&72.56&77.39&
			&79.10&72.56&77.39\\
			\bottomrule	
			
		\end{tabular}
	}
	\label{tab:1-2}
\end{table*}

\subsection{Comparison to State-of-the-Art Methods}
In this section, we evaluate the CSS performance of our proposed method on Pascal VOC and ADE20k datasets, against existing state-of-the-art methods, including PI~\cite{zenke2017continual}, EWC~\cite{kirkpatrick2017overcoming}, RW~\cite{chaudhry2018riemannian},  LwF~\cite{li2017learning}, LwF-MC~\cite{rebuffi2017icarl}, ILT~\cite{michieli2019incremental}, MiB~\cite{cermelli2020modeling}
and PLOP~\cite{douillard2021plop}. In the tables, we also provide the results of the other two methods: the simple fine-tuning approach which trains the model on the new images with no additional constraints (denoted by ``FT"), and training the model on all classes off-line (denoted by ``Joint"). The former can be regarded as a lower limit and the latter as an upper limit.

Table~\ref{tab:1-2} summarizes the experimental results for the \emph{Disjoint} and \emph{Overlapped} settings of three VOC benchmarks. Under the \emph{Disjoint} setting, it is observed that the performance of our method consistently surpasses the other methods at the last learning step on each evaluated benchmark. On VOC-19-1, we can see that the mIOU of our method on new classes (20) is $6.90\%$ higher than that of PLOP. On the VOC-15-1 with a large number of learning steps, our method consistently performs better than other methods. All of these results indicate the effectiveness of our method to catastrophic forgetting of past classes and overfitting on the current classes. Under the \emph{Overlapped} setting, we can see that the performance of our method consistently outperforms that of other methods by a sizable margin on all evaluated VOC benchmarks (i.e., 19-1, 15-5 and 15-1). 
On VOC-19-1, the forgetting of old classes (1-19) is reduced by $1.91\%$ while performance on new classes is greatly improved by $18.25\%$. On the most challenging benchmark VOC-15-1, it is worth noting that the performance of our method on all the seen classes outperforms its closest contender PLOP~\cite{douillard2021plop} by around $7.48\%$. The overview on the performance of new classes reveals that our approach is greatly helpful to avoid the overfitting on new classes while maintaining the performance on previous classes.

%% file: sections/conclusion.tex
% !TeX spellcheck = en_US
\section{Conclusion and Limitation}
In this paper, we first consider the biased context problem in CSS and design a novel biased-context-rectified CSS framework for it. 
Firstly, our method utilizes a context-rectified image-duplet learning scheme and a biased-context-insensitive consistency loss to rectify the biased context correlation between the old-class pixels and new-class pixels, which effectively alleviates the old-class forgetting and new-class overfitting. Secondly, we propose an adaptive re-weighting class-balanced learning strategy to cope with the dynamiclly changing imbalanced class distribution in CSS scenario. 
Lastly, we perform intensive evaluations of our method and other SOTA CSS methods, showing the effectiveness of our method.